\documentclass{article}
\usepackage{ijcai26}

\usepackage{times}
\usepackage{soul}
\usepackage{url}
\usepackage[hidelinks]{hyperref}
\usepackage[utf8]{inputenc}
\usepackage[small]{caption}
\usepackage{graphicx}
\usepackage{amsmath}
\usepackage{amsthm}
\usepackage{booktabs}
\usepackage{algorithm}
\usepackage{algorithmic}
\usepackage[switch]{lineno}
\usepackage{amssymb}
\usepackage{textcomp}
\usepackage{stfloats}
\usepackage{verbatim}
\usepackage{multirow}
\usepackage{pifont,makecell}
\usepackage{xcolor}
\usepackage{threeparttable}
\usepackage{pifont}
\newcommand{\cmark}{\ding{51}} 
\newcommand{\xmark}{\ding{55}} 

\title{Decoupling Perception and Calibration: Label-Efficient Image Quality Assessment Framework}

\author{
Xinyue Li$^1$
\and
Zhichao Zhang$^1$
\and
Zhiming Xu$^2$\and
Shubo Xu$^3$\and
Xiongkuo Min$^1$\and
Yitong Chen$^1$$^{\dagger}$\And
Guangtao Zhai$^1$$^{\dagger}$\\
\affiliations
$^1$Shanghai Jiao Tong University,
$^2$Xi'an Jiaotong University,
$^3$Baidu\\
\emails
\{xinyueli, liquortect\}@sjtu.edu.cn,
xzm060320@stu.xjtu.edu.cn,
xushubo@baidu.com,
\{minxiongkuo,yitongchen,zhaiguangtao\}@sjtu.edu.cn
}

\begin{document}

\maketitle

\begin{abstract}

Recent multimodal large language models (MLLMs) have demonstrated strong capabilities in image quality assessment (IQA) tasks. However, adapting such large-scale models is computationally expensive and still relies on substantial Mean Opinion Score (MOS) annotations. We argue that for MLLM-based IQA, the core bottleneck lies not in the quality perception capacity of MLLMs, but in MOS scale calibration. Therefore, we propose LEAF, a \textit{\textbf{\underline{L}}abel-\textbf{\underline{E}}fficient Image Quality \textbf{\underline{A}}ssessment \textbf{\underline{F}}ramework} that distills perceptual quality priors from an MLLM teacher into a lightweight student regressor, enabling MOS calibration with minimal human supervision. Specifically, the teacher conducts dense supervision through point-wise judgments and pair-wise preferences, with an estimate of decision reliability. Guided by these signals, the student learns the teacher’s quality perception patterns through joint distillation and is calibrated on a small MOS subset to align with human annotations. Experiments on both user-generated and AI-generated IQA benchmarks demonstrate that our method significantly reduces the need for human annotations while maintaining strong MOS-aligned correlations, making lightweight IQA practical under limited annotation budgets. The code will be released upon the publication.

\end{abstract}

\section{Introduction}
\label{sec_intro}

Image Quality Assessment (IQA) is a key component of modern visual systems, supporting a wide range of applications including camera calibration, image enhancement, content filtering, streaming media optimization, and large-scale data management~\cite{spaq,HTTP-QA-survey}.
IQA facilitates efficient benchmarking and optimization of image enhancement and generative models by providing feedback that is consistent with human standards~\cite{DISTS,AGHIQA,LIQE,CLIP_IQA}.

\begin{figure}
    \centering
    \includegraphics[width=1\linewidth]{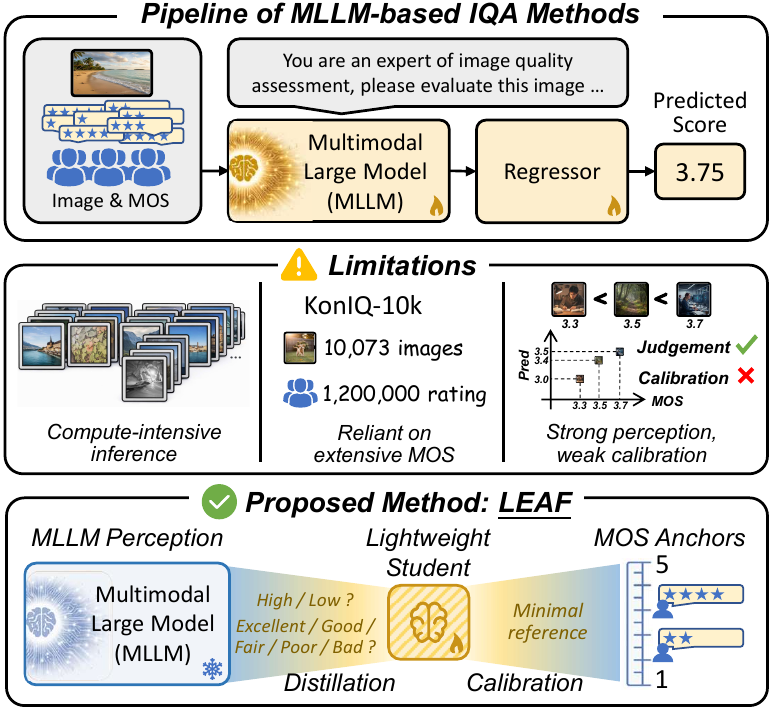}
    \caption{Using MLLMs for IQA is computationally expensive and annotation-intensive. MLLMs can perceive the key information of image quality. This work advocates decoupling quality perception from MOS calibration, thereby enabling efficient image quality assessment with minimal human supervision.}
    \label{fig_1}
\end{figure}

Multimodal large language models (MLLMs) have recently demonstrated outstanding capabilities in IQA tasks, often matching specialized IQA models by leveraging strong visual understanding and rich prior knowledge~\cite{Zhang_CVPR,MA_AGIQA,MLLM-IQA}.
This advancement has led to an emerging practice, treating MLLMs as high-performing quality evaluators and adapting them for IQA through instruction tuning or task-specific fine-tuning~\cite{AGHIQA,LMM4LMM,SEAGULL}.
However, despite the considerable accuracy of these MLLM-based methods, they still face two practical barriers that limit their usability in real-world IQA pipelines, as shown in Figure~\ref{fig_1}.

First, MLLM-based methods are computationally expensive throughout the pipeline, both during training and at deployment\cite{MLLM-IQA}. For example, ~\cite{LMM4LMM} and ~\cite{LMM4Edit} proposed MLLM-based IQA methods, which require training and inference on large memory GPUs. This makes large MLLMs difficult to use in settings that demand fast, low-cost, and scalable quality prediction, such as on-device assessment, large-scale data filtering, and real-time monitoring\cite{MLLM-IQA}.

Second, current IQA methods mainly rely on supervised learning with human mean opinion scores (MOS)~\cite{AGHIQA,MA_AGIQA}. Although MOS can provide reliable subjective judgments under controlled protocols, MOS annotation is resource-intensive. For example, the KonIQ-10K~\cite{KonIQ10k} dataset contains more than $10,000$ images and has collected a total of $1.2$ million opinion scores. Moreover, subjective quality perception drifts with the scenario. Therefore, MOS annotations need to be recollected as scenarios and time changes.

\begin{figure*}
  \centering
  \includegraphics[width=0.33\linewidth]{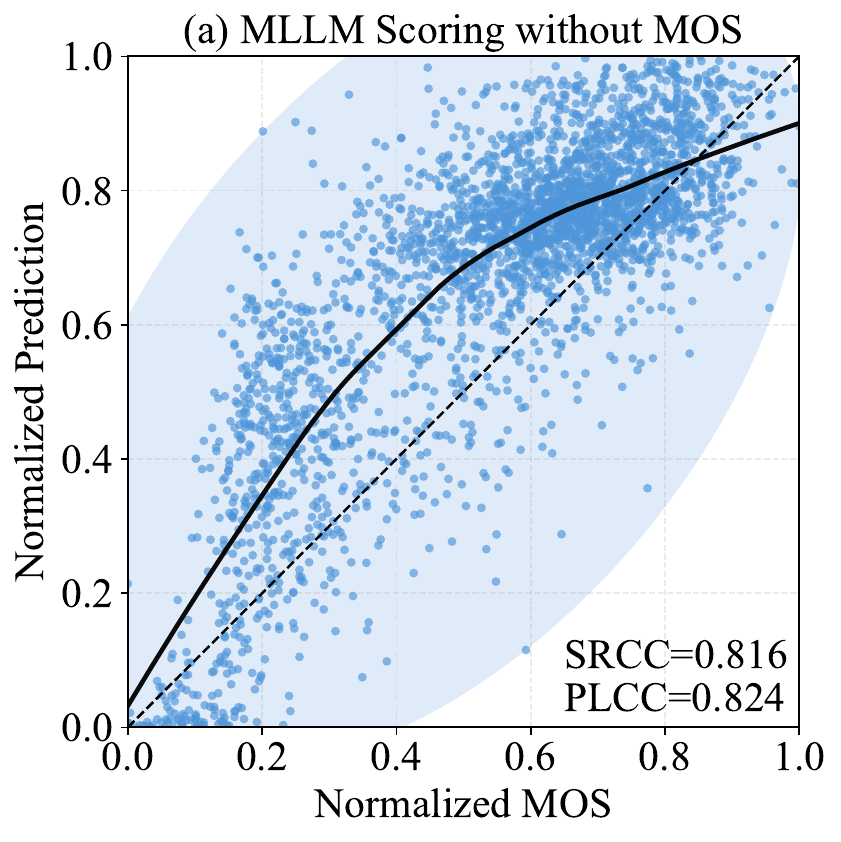}
  \includegraphics[width=0.33\linewidth]{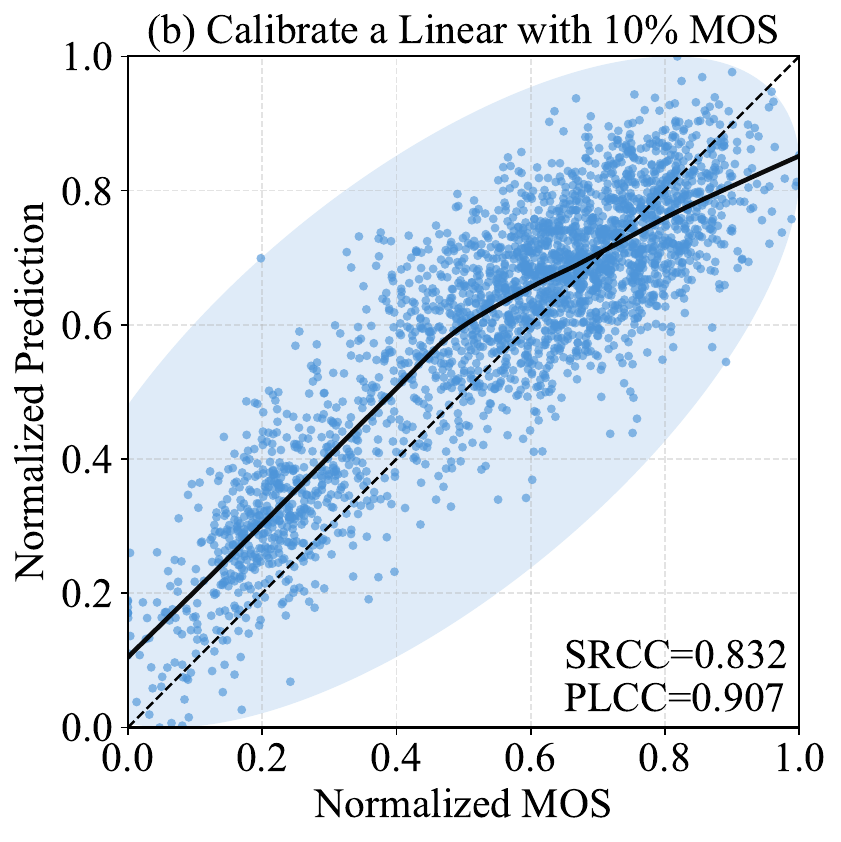}
  \includegraphics[width=0.33\linewidth]{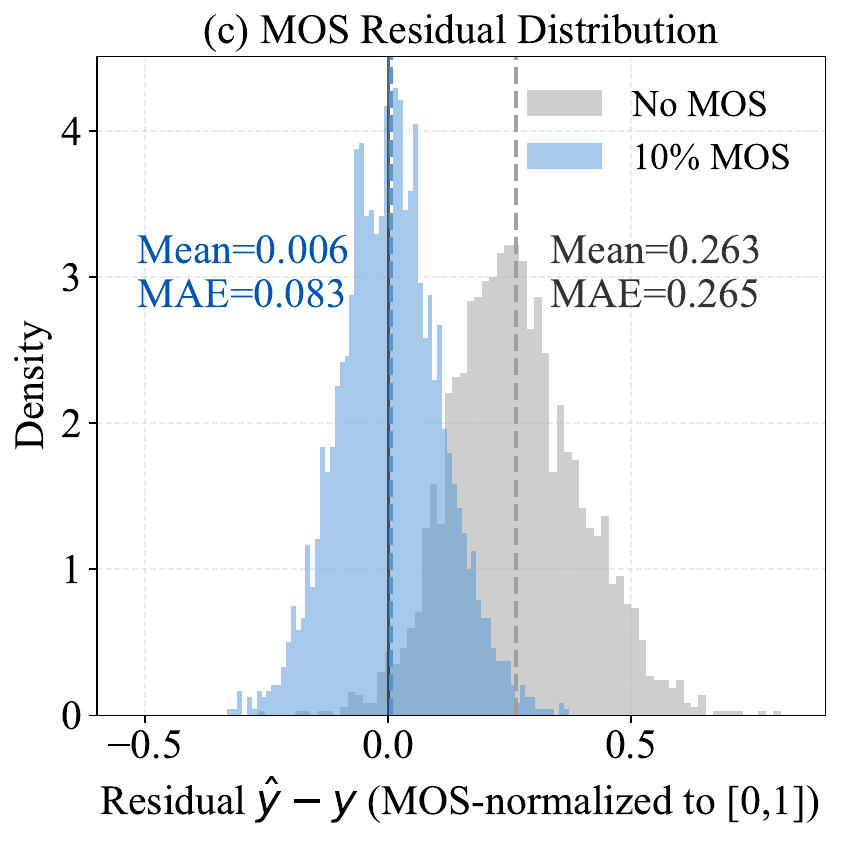}
  \caption{Direct MLLM scoring on AGIQA-3K exhibits strong monotonic correlation with MOS (SRCC=$0.816$) but suffers from pronounced scale bias and non-linear mapping (shown in (a)). Calibrating a Linear with only $10\%$ MOS substantially improves MOS alignment (PLCC from $0.824$ to $0.907$) and reduces residual bias (mean residual from $0.263$ to $0.006$).}
  \label{fig_2}
\end{figure*}

More importantly, for a pretrained MLLM, the primary limitation is often not the capability of quality perception, but the absence of explicit MOS scale calibration. 
Specifically, MLLMs perform well in capturing comparative and quality classification, such as distinguishing better versus worse images or assigning coarse quality levels. However, they struggle to map these perceptual judgments onto the MOS rating scales of specific datasets and specific scenarios.


Figure~\ref{fig_2} illustrates this phenomenon on the AGIQA-3K dataset. Direct MLLM scoring preserves reliable ranking, but suffers from clear MOS scale miscalibration. Calibrating only a lightweight head with $10\%$ MOS largely removes the residual bias and improves MOS alignment while keeping ranking performance (Fig.~\ref{fig_2}(b,c)). This suggests that the key challenge is MOS scale calibration rather than relearning perception. Consequently, directly fine-tuning MLLMs using large-scale MOS annotations is an inefficient strategy. More experimental details are reported in the supplementary materials.

Inspired by this observation, in this work, we answer this question with \textbf{LEAF}, a \textit{\textbf{\underline{L}}abel-\textbf{\underline{E}}fficient Image Quality \textbf{\underline{A}}ssessment \textbf{\underline{F}}ramework} that transfers perceptual quality knowledge from an MLLM teacher into a lightweight student.
The key idea is to use the teacher to provide dense supervision on a large unlabeled image set, and reserve human MOS labels only for final calibration.

Concretely, the teacher provides two complementary forms of supervision:
(i) \textbf{point-wise} quality judgments for individual images, and
(ii) \textbf{pair-wise} preferences that encode relative quality rankings, and also include estimates of decision reliability. 
Leveraging these signals, we formulate a Joint Teacher-Guided Distillation to enable the student to learn the teacher’s quality perception.
We then conduct Calibration Fine-Tuning on only a small MOS subset to align the student's scoring scale with human judgments.
 
Our contributions are summarized as follows:
\begin{itemize}
\item We formulate a label-efficient IQA framework that decouples perceptual knowledge from MOS calibration, based on the insight that MLLMs are strong at quality perception but weak at dataset-specific MOS alignment.
\item We propose a two-stage teacher-guided framework that transfers  perceptual quality priors to a lightweight student by joint distillation of point-wise judgments and confidence-weighted pair-wise preferences, followed by calibration using a small MOS-labeled subset.
\item Experiments on both UGC and AIGC benchmarks validate the proposed decoupling principle, showing that strong MOS-aligned correlation can be achieved with limited MOS supervision, while avoiding costly task-specific fine-tuning of MLLMs.
\end{itemize}

\section{Related Work}
\label{sec_related_works}

\subsection{Image Quality Assessment}

\subsubsection{Handcrafted IQA}
Early image quality assessment (IQA) relied on handcrafted fidelity and perception priors, among which SSIM-style structural comparison methods became influential and were further extended to improve robustness under diverse distortions~\cite{MS_SSIM}. Complementary handcrafted cues were explored through phase and gradient features to better capture perceptually significant structural changes~\cite{FSIM}. No-reference IQA subsequently adopted natural scene statistics to treat distortions as a deviation from natural regularities, as demonstrated by DIIVINE~\cite{DIIVINE}, BRISQUE~\cite{BRISQUE}, and the opinion-unaware NIQE~\cite{NIQE}.

\subsubsection{Deep Learning--based IQA}
Deep IQA learns quality-aware representations directly from data, while WaDIQaM~\cite{WaDIQaM} is an early blind image quality assessment model based on CNN. Full-reference IQA also shifted to deep feature distances, such as LPIPS~\cite{LPIPS}, DISTS~\cite{DISTS}, and PieAPP~\cite{PieAPP}. Later blind IQA methods improved adaptability and spatial sensitivity through hypernetworks in HyperIQA~\cite{HyperIQA}, attention mechanisms~\cite{NRIQA}, pseudo-reference hallucination in HIQA~\cite{HIQA}, and stronger CNN interactions~\cite{BIQA}, while transformers such as MUSIQ~\cite{MUSIQ} and MANIQA~\cite{MANIQA} further strengthen global and multi-scale modeling.

\subsubsection{Multi-modal Model-based IQA}
Multimodal IQA exploits vision-language priors, where LIQE~\cite{LIQE} combines image and text embeddings and CLIP-IQA~\cite{CLIP_IQA} supports prompt-based zero-shot assessment. For AI-generated images, prompt consistency can be explicitly modeled within the CLIP-based frameworks, and multimodal large models can also provide discrete judgments and continuous scoring for quality assessment~\cite{CLIP_AGIQA,Zhang_CVPR}. Recent AIGC-IQA frameworks include MA-AGIQA~\cite{MA_AGIQA} and SEAGULL~\cite{SEAGULL}, while EvoQuality~\cite{EvoQuality} explore zero-shot or self-evolving ranking strategies, and AGHI-QA~\cite{AGHIQA} provides a dedicated benchmark, but most methods still rely on annotation and motivate label-free alternatives.

\subsection{Label-Free IQA Methods}
\subsubsection{Classical NSS-based Model}
Although supervised IQA achieves strong performance, collecting large-scale MOS labels is costly and difficult to scale to new domains and rapidly evolving AI-generated content. This motivates the emergence of label-free IQA, which estimates perceptual quality without explicit MOS supervision. Early methods follow the NIQE-style paradigm, they model pristine natural-scene statistics and quantify quality by the deviations from learned priors, e.g., NIQE~\cite{NIQE}, IL-NIQE~\cite{ILNIQE}, and uBIQA~\cite{uBIQA}. This idea has been further extended to domain-specific scenarios such as medical IQA (MSM-MIQ~\cite{MSMMIQA}).

\subsubsection{Deep Distribution Model}
Beyond the hand-crafted NSS method, deep distribution modeling aims to better characterize high-quality image priors. MDFS~\cite{MDFS} learns multi-scale deep feature distributions, while DSTS~\cite{DSTS} separates shape- and texture-biased representations to capture various degradation scenarios. Generative and self-supervised variants learn quality-related latent spaces or features for deviation-based scoring, such as adversarial CVAE-based opinion-unaware perceptual IQA~\cite{AdvCVAEIQA} and NROUQA~\cite{NROUQA}. Despite different implementation approaches, these methods remain largely consistent with the NIQE-style principle of deviation from the original state.

\subsubsection{Pretraining and Quality Representation}
Another line focuses on achieving label-free quality-aware representation learning through self-supervised or pretext-based objectives. QPT~\cite{QaPMBIQA} introduced quality-aware contrastive pretraining, ARNIQA~\cite{ARNIQA}, and CONTRIQUE~\cite{CONTRIQUE} utilized synthetic degradation sequences and distortion-conditioned contrastive learning. Transformer-based designs further combine contrastive learning with quality regression~\cite{TNRIQA}, while Re-IQA~\cite{REIQA} used a dual encoder to separate content and quality. DUBMA~\cite{DOUBIQA} leverages a large distortion bank and FR-IQA evaluation metrics as pseudo-annotators for ranking supervision, and QualiCLIP~\cite{QualiCLIP} aligned degraded images with quality-related antonymic prompts to learn multimodal embeddings while avoiding the use of MOS.

\begin{figure*}
    \centering
    \includegraphics[width=1\linewidth]{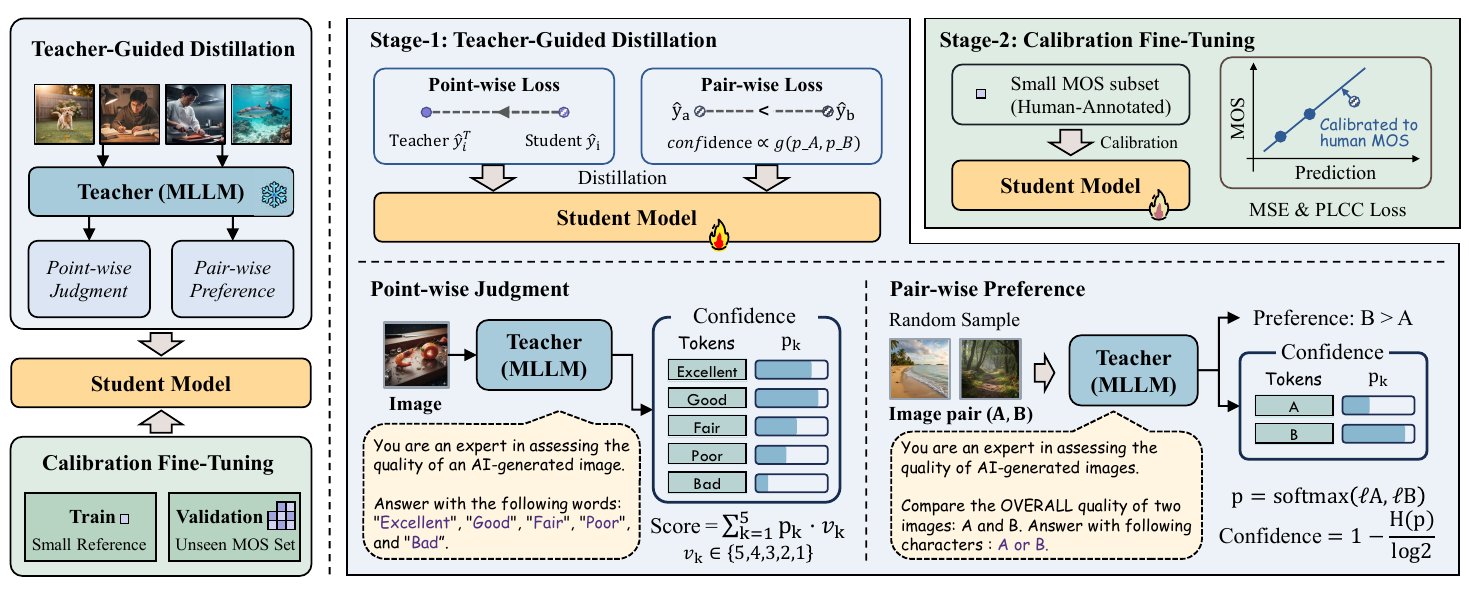}
    \caption{Overview of LEAF. Label-efficient IQA framework: Stage-1 jointly distills point-wise judgments and pair-wise preferences from an MLLM teacher into a lightweight student. Stage-2 calibrates the student to human MOS using a small annotated subset.}
    \label{fig_framework}
\end{figure*}

\section{Method}

\subsection{Overview}

To reduce the reliance on large-scale human MOS annotations while maintaining MOS-aligned IQA performance,
we propose \textbf{LEAF}, a \textit{\textbf{\underline{L}}abel-\textbf{\underline{E}}fficient Image Quality \textbf{\underline{A}}ssessment \textbf{\underline{F}}ramework} with two components:
(i) \emph{Teacher-Guided Distillation} that produces point-wise judgment and pair-wise preferences, and the student learns from dense teacher signals without using MOS in the distillation objective at stage 1,
and (ii) \emph{Calibration Fine-Tuning} that calibrates the student model using a small MOS subset at stage 2.
An overview is shown in Figure~\ref{fig_framework}.


Let $\mathcal{D}=\{x_i\}_{i=1}^{N}$ be a set of training images, and let only a small subset $\mathcal{D}_{\mathrm{MOS}}=\{(x_i,y_i)\}_{i=1}^{M}\subset\mathcal{D}$ be annotated with human MOS $y_i\in\mathbb{R}$, where $M\ll N$.
Our goal is to learn a lightweight student regressor $s_\theta:\mathcal{X}\rightarrow\mathbb{R}$ that predicts perceptual quality scores aligned with MOS.

\subsubsection{Teacher-Guided Distillation}
Given an image $x\in\mathcal{D}$, the MLLM teacher provides:
(i) a point-wise judgment $\hat{y}^{T}(x)$, and
(ii) a set of pair-wise supervision tuples
$\mathcal{P}=\{(x_a,x_b,t_{ab},\omega_{ab})\}$ with $(x_a,x_b)\in\mathcal{D}\times\mathcal{D}$.
We construct $\mathcal{P}$ by uniformly sampling image pairs at random from $\mathcal{D}\times\mathcal{D}$.
For each pair, we assign $x_a$ as option A and $x_b$ as option B in the teacher prompt.
Here $t_{ab}\in\{0,1\}$ indicates the teacher's preference, where $t_{ab}=1$ if $x_a$ is judged to have higher quality than $x_b$,
and $\omega_{ab}\in[0,1]$ measures the confidence of this decision.

During this stage, student $s_\theta$ is trained through dense teacher-induced supervision, without using MOS labels in the distillation objective.
Concretely, the student is optimized to (i) match the teacher's point-wise judgment $\hat{y}^{T}(x)$ and
(ii) preserve the pairwise ranking induced by the teacher, weighted by the teacher's confidence encoded in $\mathcal{P}$.

\subsubsection{Calibration Fine-Tuning} 
This stage uses the small MOS-labeled subset $\mathcal{D}_{\mathrm{MOS}}$ to calibrate the student predictions to the MOS scale.
We fine-tune the student on $\mathcal{D}_{\mathrm{MOS}}$ with regression objective together with the correlation-based objective between predictions $\{s_\theta(x_i)\}$ and MOS labels $\{y_i\}$,
so that the final scores are aligned with human perceptual judgments.

\subsection{Teacher-Guided Distillation (TGD)}
\label{sec:teacher_supervision}
The teacher provides two complementary supervision signals:
(1) \emph{point-wise} quality judgments for individual images, and
(2) \emph{pair-wise} preferences with confidence for image pairs.
We use fixed prompt templates to ensure constrained outputs. The complete prompts are provided in the appendix.

\subsubsection{Point-wise Judgment}
\label{sec:point-wise_teacher}
For each image $x$, we prompt the teacher to judge its overall perceptual quality using a fixed set of discrete quality tokens
$\mathcal{K}=\{Excellent, Good, Fair, Poor, Bad\}$.
Instead of using the teacher's hard output, we extract token-level log-likelihoods $\{\ell_k(x)\}_{k\in\mathcal{K}}$ and convert them into a probability distribution:
\begin{equation}
p_k(x) = \frac{\exp(\ell_k(x))}{\sum_{j\in\mathcal{K}}\exp(\ell_j(x))}.
\label{eq:teacher_point_prob}
\end{equation}
We then map the discrete tokens to ordinal scores
$\{v_k\}=\{5,4,3,2,1\}$ (from ${Excellent}$ to ${Bad}$)
and define a continuous teacher score by expectation:
\begin{equation}
\hat{y}^{T}(x) = \sum_{k\in\mathcal{K}} v_k\, p_k(x).
\label{eq:teacher_point_score}
\end{equation}
This produces a dense supervision signal for all images in $\mathcal{D}$.

\subsubsection{Pair-wise Preference}
\label{sec:pair-wise_teacher}
To capture relative quality, we randomly sample pairs $(x_a, x_b)$ and prompt the teacher to judge which image has higher quality,
outputting a single token in $\{\texttt{A}, \texttt{B}\}$.

We obtain the log-likelihoods $(\ell_A, \ell_B)$ at the decision position and compute:
\begin{equation}
p_A=\frac{e^{\ell_A}}{e^{\ell_A}+e^{\ell_B}},\quad
p_B=\frac{e^{\ell_B}}{e^{\ell_A}+e^{\ell_B}}.
\label{eq:pair_prob}
\end{equation}
The hard preference label is
\begin{equation}
t_{ab} =
\begin{cases}
1, & \text{if teacher prefers } A \ (p_A \ge p_B),\\
0, & \text{otherwise}.
\end{cases}
\label{eq:pref_label}
\end{equation}

Not all teacher pair-wise decisions are equally reliable.
We quantify confidence through the normalized entropy of the two-class distribution: 
\begin{equation}
H(p) = -\sum_{c\in\{A,B\}} p_c \log p_c,\qquad
\omega_{ab} = 1 - \frac{H(p)}{\log 2}\in[0,1],
\label{eq:conf_weight_entropy}
\end{equation}
where $\omega_{ab}$ is larger when teacher's decision is more certain.
In practice, we discard low-confidence pairs with $\omega_{ab}<\tau$.

\subsubsection{Distillation Objective}

We distill quality prior from the teacher using dense supervision on $\mathcal{D}$.
Specifically, we leverage both (i) pointwise soft scores $\hat{y}^{T}(x)$ and (ii) pairwise preferences $\mathcal{P}$ to train a lightweight student regressor.

\paragraph{Student predictor.}
We denote the student as an IQA regressor $s_\theta:\mathcal{X}\rightarrow\mathbb{R}$ that outputs a scalar quality score.
Our formulation does not rely on a specific architecture, requiring only real-valued outputs suitable for regression and ranking.

\paragraph{Point-wise distillation.}
Given teacher-provided soft scores $\hat{y}^{T}(x)$ for $x\sim\mathcal{D}$, we minimize
\begin{equation}
\mathcal{L}_{\text{reg}} =
\mathbb{E}_{x\sim \mathcal{D}}
\left[
\operatorname{SmoothL1}\big(s_\theta(x), \hat{y}^{T}(x)\big)
\right].
\label{eq:stage1_reg}
\end{equation}

\paragraph{Pair-wise distillation.}
Given teacher-provided pairs $\mathcal{P}=\{(x_a,x_b,t_{ab},\omega_{ab})\}$, we define
\begin{equation}
P_\theta(a\succ b)=\sigma\big(s_\theta(x_a)-s_\theta(x_b)\big),
\qquad
\sigma(z)=\frac{1}{1+e^{-z}}.
\label{eq:rank_prob}
\end{equation}
We optimize a weighted binary cross-entropy loss:
\begin{equation}
\mathcal{L}_{\text{rank}} =
\mathbb{E}_{(x_a,x_b,t_{ab},\omega_{ab})\sim\mathcal{P}}
\left[
\omega_{ab}\cdot
\operatorname{BCE}\big(P_\theta(a\succ b),\, t_{ab}\big)
\right].
\label{eq:stage1_rank}
\end{equation}

\paragraph{Joint objective.}
The objective of this stage is
\begin{equation}
\mathcal{L}_{\text{dis}} = \mathcal{L}_{\text{reg}} + \lambda_{dis} \mathcal{L}_{\text{rank}},
\label{eq:stage1_total}
\end{equation}
where $\lambda_{dis}$ is used to balance point-wise and pair-wise distillation.
We select the checkpoint without using any MOS annotations in the Mos free setting. In the few-shot setting, checkpoint selection is performed using a held-out split of the same MOS subset $\mathcal{D}_{\mathrm{MOS}}$ that is visible for calibration fine-tuning stage, without introducing any additional labeled data or test-set leakage.

\subsection{Calibration Fine-Tuning (CFT)}
\label{sec:stage2}

To align the student predictions with the human MOS scale, we fine-tune the student on the small labeled subset
$\mathcal{D}_{\text{MOS}}$ with a regression and a correlation objective.

Let $\mathbf{s}=[s_\theta(x_i)]_{(x_i,y_i)\in\mathcal{D}_{\text{MOS}}}$ and $\mathbf{y}=[y_i]$ denote vectors of predictions and MOS.
We first perform scale calibration using mean squared error:
\begin{equation}
\mathcal{L}_{\text{MSE}} =
\mathbb{E}_{(x_i,y_i)\sim \mathcal{D}_{\text{MOS}}}
\left[
\big(s_\theta(x_i)-y_i\big)^2
\right].
\label{eq:mse_loss}
\end{equation}
Meanwhile, we optimize the PLCC calculated within each mini-batch for differentiability by minimizing

\begin{equation}
\mathcal{L}_{\text{PLCC}} = 1 - \rho(\mathbf{s}, \mathbf{y}), \qquad
\rho(\mathbf{s},\mathbf{y}) =
\frac{\operatorname{cov}(\mathbf{s},\mathbf{y})}
{\sigma(\mathbf{s})\,\sigma(\mathbf{y})}.
\label{eq:plcc_loss}
\end{equation}
The overall objective of the calibration fine-tuning stage is
\begin{equation}
\mathcal{L}_{\text{cal}} = \mathcal{L}_{\text{MSE}} + \lambda_{cal} \mathcal{L}_{\text{PLCC}},
\label{eq:stage2_total}
\end{equation}
where $\lambda_{cal}$ balances absolute MOS calibration and correlation optimization.

\section{Experiments}

\begin{table}[t]
\caption{Comparison with state-of-the-art IQA methods on AIGC benchmarks.}
\label{tab_performance_aigc}
\centering
\small
\setlength{\tabcolsep}{4pt}   
\renewcommand{\arraystretch}{1.0} 
\begin{tabular}{l cc cc}
\toprule
Methods &
\multicolumn{2}{c}{AGIQA-3K} &
\multicolumn{2}{c}{AIGIQA-20K} \\
\cmidrule(lr){2-3}\cmidrule(lr){4-5}
 & SRCC & PLCC & SRCC & PLCC \\
\midrule

\multicolumn{5}{c}{\textbf{\textit{Supervised}}} \\

BRISQUE (TIP, 2012)          & 0.472 & 0.561 & 0.466 & 0.558 \\
HyperIQA (CVPR, 2020)        & 0.850 & 0.904 & 0.816 & 0.832 \\
MANIQA (CVPR, 2022)          & 0.861 & 0.911 & 0.850 & 0.887 \\
DBCNN (TCSVT, 2020)          & 0.826 & 0.890 & 0.805 & 0.848 \\
MUSIQ (ICCV, 2021)           & 0.820 & 0.865 & 0.832 & 0.864 \\
StairIQA (JSTSP, 2023)       & 0.834 & 0.893 & 0.789 & 0.842 \\
Q-Align (TOMM, 2023)         & 0.852 & 0.881 & 0.874 & 0.889 \\
MA-AGIQA (ACMMM, 2024)       & 0.893 & 0.927 & 0.864 & 0.905 \\

\midrule
\multicolumn{5}{c}{\textbf{\textit{Weak-supervised}}} \\

CONTRIQUE (TIP, 2022)        & 0.817 & 0.879 & 0.788 & 0.807 \\
Re-IQA (CVPR, 2023)          & 0.811 & 0.874 & 0.787 & 0.811 \\
CLIP-IQA+ (AAAI, 2023)       & \underline{0.844} & 0.894 & 0.833 & 0.854 \\
ARNIQA (WACV, 2024)          & 0.803 & 0.881 & 0.778 & 0.792 \\
GRepQ-D (WACV, 2024)         & 0.807 & 0.858 & 0.789 & 0.810 \\
\textbf{Ours (10\% MOS)}     & 0.841 & \underline{0.899} & \underline{0.839} & \underline{0.878} \\
\textbf{Ours (30\% MOS)}     & \textbf{0.868} & \textbf{0.914} & \textbf{0.860} & \textbf{0.905} \\

\midrule
\multicolumn{5}{c}{\textbf{\textit{Label-free}}} \\

NIQE (ISPL, 2012)            & 0.523 & 0.566 & 0.208 & 0.337 \\
ILNIQE (TIP, 2015)           & 0.609 & 0.655 & 0.335 & 0.455 \\
CLIP-IQA (AAAI, 2023)        & 0.638 & 0.711 & 0.388 & 0.537 \\
MDFS (TMM, 2024)             & 0.672 & 0.676 & 0.691 & 0.695 \\
QualiCLIP (ArXiv, 2025)      & 0.667 & \underline{0.735} & 0.679 & 0.694 \\
GRepQ-Z (WACV, 2024)         & 0.613 & 0.734 & 0.624 & 0.634 \\
DUBMA (IJCAI, 2025)          & \underline{0.684} & 0.701 & \underline{0.695} & \underline{0.697} \\
\textbf{Ours}                & \textbf{0.749} & \textbf{0.811} & \textbf{0.696} & \textbf{0.762} \\

\bottomrule
\end{tabular}
\end{table}

\begin{table}[t]
\caption{Comparison with state-of-the-art weak-supervised and label-free IQA methods on UGC benchmarks.}
\label{tab_performance_ugc}
\centering
\small
\setlength{\tabcolsep}{3pt}   
\renewcommand{\arraystretch}{1.0} 
\begin{tabular}{l cc cc}
\toprule
Methods &
\multicolumn{2}{c}{KonIQ-10k} &
\multicolumn{2}{c}{SPAQ} \\
\cmidrule(lr){2-3}\cmidrule(lr){4-5}
 & SRCC & PLCC & SRCC & PLCC \\

\midrule
\multicolumn{5}{c}{\textbf{\textit{Weak-supervised}}} \\

CONTRIQUE (TIP, 2022)        & 0.894 & 0.906 & \underline{0.914} & 0.919 \\
CLIP-IQA+ (AAAI, 2023)       & \underline{0.895} & \underline{0.909} & 0.864 & 0.866 \\
ARNIQA (WACV, 2024)          & 0.869 & 0.883 & 0.904 & 0.909 \\
GRepQ-D (WACV, 2024)         & 0.855 & 0.868 & 0.903 & \underline{0.917} \\
\textbf{Ours (10\% MOS)}        & 0.867 & 0.903 & 0.896 & 0.902 \\
\textbf{Ours (30\% MOS)}        & \textbf{0.899} & \textbf{0.916} & \textbf{0.921} & \textbf{0.922} \\

\midrule
\multicolumn{5}{c}{\textbf{\textit{Label-free}}} \\

NIQE (ISPL, 2012)            & 0.551 & 0.488 & 0.703 & 0.670 \\
ILNIQE (TIP, 2015)           & 0.453 & 0.467 & 0.719 & 0.654 \\
CONTRIQUE (TIP, 2022)                  & 0.651 & 0.637 & 0.677 & 0.685 \\
CL-MI  (WACV, 2023)                      & 0.664 & 0.653 & 0.701 & 0.701 \\
Re-IQA (CVPR, 2023)                & 0.580 & 0.568 & 0.613 & 0.616 \\
CLIP-IQA (AAAI, 2023)        & 0.695 & 0.727 & 0.738 & 0.735 \\
ARNIQA (WACV, 2024)           & 0.746 & 0.762 & 0.788 & 0.797 \\
MDFS (TMM, 2024)             & 0.733 & 0.737 & 0.741 & 0.754 \\
GRepQ-Z (WACV, 2024)          & \underline{0.768} & \underline{0.784} & 0.823 & 0.839 \\
DUBMA (IJCAI, 2025)          & 0.703 & 0.740 & \underline{0.834} & \underline{0.841} \\
\textbf{Ours}                & \textbf{0.777} & \textbf{0.801} & \textbf{0.861} & \textbf{0.867} \\

\bottomrule
\end{tabular}
\end{table}

\subsection{Experimental Setup}
\label{sec:exp_setup}
The teacher model is based on the InternVL-3.5-8B and remains frozen during all stages. 
We use a ConvNeXt-Base as the student model pretrained on ImageNet. 
All images are resized to 256 and cropped to $224\times224$. Random resized cropping and horizontal flipping are applied during training, while center cropping is used for evaluation. 
We use AdamW for model optimization and adopt mixed precision training. 
Stage 1 is trained jointly with the objective by using point-wise judgments and pairwise preferences provided by the teacher, where $\lambda_{dis} = 0.5$.
Stage-2 fine-tunes the student on the labeled subset using a correlation-based objective to align the prediction results with  human MOS.
Hyperparameters and implementation details, including pair construction and filtering strategies, are provided in the supplementary material.

\subsection{Datasets and Evaluation Protocol}
\label{sec:exp_datasets}

We evaluate on the image quality assessment benchmarks for both user-generated content (UGC) and AI-generated content (AIGC). 
For UGC, we use KonIQ-10k~\cite{KonIQ10k} and SPAQ~\cite{spaq}, which contain natural images with authentic distortions from real-world capture and processing pipelines. 
For AIGC, we use AGIQA-3K~\cite{AGIQA3K}, AIGIQA-20K~\cite{AIGIQA20K}, which cover images generated by diverse text-to-image models with varying perceptual qualities. 

To quantify label efficiency, we adjusted the MOS-visible ratio on the training split, resulting in three evaluation settings that correspond to Table~\ref{tab_performance_aigc} and Table~\ref{tab_performance_ugc}.

(i) label-free: the student is trained solely with teacher supervision (e.g., predicted scores and preferences), without using any human MOS.

(ii) $10\%$ MOS: only a randomly sampled 10\% subset of training images contain MOS labels, while the remaining training samples are treated as MOS-invisible.

(iii) $30\%$ MOS: similarly, 30\% of the training images contain MOS labels.

We repeat the sampling process five times with different random seeds and report the average SRCC/PLCC values across the five runs.

Across all settings, we construct the pair-wise supervision set by uniformly sampling image pairs at random, with the number of sampled pairs matching the size of the dataset. 
The ablation experiment on the number of samples is provided in the supplementary material.

\paragraph{Metrics.}
We report image quality assessment metrics based on standard correlation, including SRCC and PLCC, computed on the held-out test split.
Relevant definitions and specific implementation details can be found in the supplementary materials. 

\subsection{Comparison with State-of-the-Art Methods}
We compare our method with representative weak-supervised and label-free IQA approaches on both AIGC and UGC benchmarks. For AIGC, we report results on AGIQA-3K and AIGIQA-20K in Table~\ref{tab_performance_aigc}. For UGC, we evaluate the performance on KonIQ-10k and SPAQ in Table~\ref{tab_performance_ugc}. In each case, we consider both label-free and weak-supervised settings to validate the effectiveness of the proposed method.

\subsubsection{AIGC Benchmarks}

Without using any human MOS, our method shows strong correlation with human judgments on AGIQA-3K, with an SRCC of $0.749$ and a PLCC of $0.811$, outperforming the strongest label-free baseline in the table (SRCC = $0.684$). On AIGIQA-20K, it reaches $0.696$ and $0.762$ in terms of SRCC and PLCC, respectively, achieving the best SRCC among label-free competitors while exceeding the best label-free PLCC $0.697$. These results indicate that teacher-driven dense supervision can transfer reliable quality perception even without human labels.

With ten percent MOS, our method achieves an SRCC of $0.841$ and a PLCC of $0.899$ on AGIQA-3K, and $0.839$ and $0.878$ on AIGIQA-20K, respectively. On AGIQA-3K, the PLCC is the highest among weak-supervised methods, while the SRCC remains close to the strongest competitor. For reference, most weak-supervised IQA methods in the table are evaluated using a large portion of MOS labels (often around $70\%$~\cite{REIQA,ARNIQA,CONTRIQUE}) to calibrate the regression head. 
In contrast, we report results with $30\%$ MOS in our framework, with the performance on AGIQA-3K further improving to $0.868$ and $0.914$, and to $0.860$ and $0.905$ on AIGIQA-20K in terms of SRCC and PLCC, respectively. Notably, the PLCC on AIGIQA-20K matches the best supervised result reported in the table.

\subsubsection{UGC Benchmarks}
Under the label-free setting, our method achieves strong correlations on both UGC benchmarks. On KonIQ-10k, it attains an SRCC of $0.777$ and a PLCC of $0.801$, while on SPAQ, it reaches $0.861$ and $0.867$ of SRCC and PLCC, respectively. On both datasets, our method achieved the best results among label-free competitors. In particular, on SPAQ, it improved SRCC from $0.834$ to $0.861$ compared to the strongest competing baseline, and in terms of PLCC, it improved from $0.841$ to $0.867$.

With $10\%$ MOS, our method achieves $0.867$ and $0.903$ on KonIQ-10k, and $0.896$ and $0.902$ on SPAQ in terms of SRCC and PLCC. For reference, the compared weak-supervised methods commonly rely on around $70\%$ of MOS labels for regression-based calibration\cite{CONTRIQUE,REIQA,ARNIQA}. Using only thirty percent MOS, our method further improves performance to $0.899$ and $0.916$ on KonIQ-10k, and to $0.921$ and $0.922$ on SPAQ, achieving the best results among weak-supervised methods reported in Table~\ref{tab_performance_ugc}. These results show that limited MOS supervision already yields strong performance, while additional MOS labels consistently improve the correlation.

\begin{figure}
    \centering
    \includegraphics[width=1\linewidth]{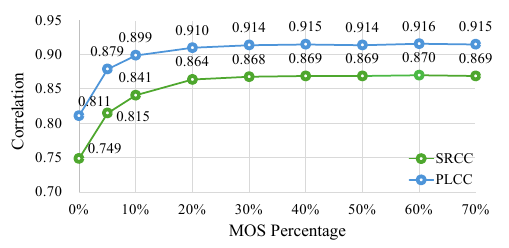}
    \caption{Impact of MOS availability during MOS calibration on AGIQA-3K.}
    \label{fig_MOS_ratio}
\end{figure}

\subsection{Label Efficiency Analysis}

Figure~\ref{fig_MOS_ratio} illustrates label efficiency on AGIQA-3K by varying the MOS availability ratio in Calibration Fine-Tuning (CFT) stage while keeping Teacher-Guided Distillation (TGD) stage fixed. Overall, performance improves sharply with a small amount of MOS and then gradually saturates. Without any MOS, the model achieves an SRCC of $0.749$ and a PLCC of $0.811$. When the MOS ratio reaches $10\%$, performance increases to $0.841$ and $0.899$ in terms of SRCC and PLCC, respectively. As the MOS ratio further increases, the gains become marginal, reaching $0.869$ and $0.915$ at $30\%$ MOS. 
These results indicate that CFT primarily serves to calibrate the dataset-specific score scale, enabling strong performance with limited MOS annotations.

\label{sec:exp_ablation}

\begin{table}[t]
\caption{Ablation study of different supervision strategies in Teacher-Guided Distillation.}
\label{tab_ablation}
\centering
\begin{tabular}{cccc|cc}
\toprule
Point & Pair & Pair Conf. &  & SRCC & PLCC \\
\midrule
\xmark & \xmark & \xmark &  & 0.759 & 0.812 \\
\cmark & \xmark & \xmark &  & 0.834 & 0.892 \\
\cmark & \cmark & \xmark &  & 0.836 & 0.895 \\
\xmark & \cmark & \cmark &  & 0.829 & 0.889 \\
\cmark & \cmark & \cmark &  & \textbf{0.841} & \textbf{0.899} \\
\bottomrule
\end{tabular}
\end{table}
\subsection{Ablation Studies}

Table~\ref{tab_ablation} presents different supervision strategies used in TGD on AGIQA-3K with $10\%$ visible MOS, including point-wise judgments, pair-wise preferences, and confidence-aware weighting on pairs. 

The first row reports a CFT-only baseline, where Stage-1 distillation is removed, resulting in $0.759$ and $0.812$.
Using only point-wise supervision to distill the teacher significantly improves the student to $0.834$ and $0.892$, indicating that point-wise soft judgments provide strong quality guidance. Adding pair-wise supervision on top of point-wise supervision further boosts performance to $0.836$ and $0.895$, showing that relative comparisons provide complementary ranking information. When relying solely on pair-wise supervision with confidence weighting, the student reaches $0.829$ and $0.889$. This setting still brings a noticeable performance gain, showing that pair-wise preferences can transfer useful relative quality information. However, the improvement is smaller than that achieved with point-wise supervision.

\subsection{Robustness to Different Backbones}
Table~\ref{tab_robustness} studies the robustness of our framework to different teacher and student choices on AGIQA-3K with $10\%$ visible MOS. Overall, stronger teachers can provide higher-quality supervision signals, and the student achieves consistent performance across various backbone network capacities.

For Qwen3-VL, SRCC and PLCC increased from $0.835$ and $0.873$ with the $8$B model to $0.848$ and $0.907$ with the $32$B model. Meanwhile, InternVL-3.5 provided a competitive and generally stronger supervision, reaching $0.841$ and $0.899$ with the $8$B model and achieving the best performance of $0.850$ and $0.912$ with the $38$B model.

For student backbones, performance improves from ResNet-18 to ConvNeXt variants, with the best performance achieved by ConvNeXt-Base with $0.841$ and $0.899$. Further expansion to ConvNeXt-Large does not bring additional gains, with results of $0.836$ and $0.882$, suggesting that a moderately sized backbone offers a better trade-off between capacity and optimization stability under the same training protocol.

\begin{table}[t]
\centering
\caption{Robustness to teacher models and student backbone architectures.}
\label{tab_robustness}
\begin{tabular}{lccc}
\toprule
Setting & Params & SRCC & PLCC \\
\midrule
\multicolumn{4}{c}{\textbf{MLLMs}} \\
\midrule
Qwen3-VL        & 8B  & 0.835 & 0.873 \\
Qwen3-VL        & 32B & 0.848 & 0.907 \\
InternVL-3.5   & 8B  & 0.841 & 0.899 \\
InternVL-3.5   & 38B & 0.850 & 0.912 \\
\midrule
\multicolumn{4}{c}{\textbf{Student Backbones}} \\
\midrule
ResNet-18        & 11.7M & 0.807 & 0.868 \\
ResNet-50        & 25.6M & 0.815 & 0.886 \\
ConvNeXt-Tiny    & 28.6M & 0.833 & 0.896 \\
ConvNeXt-Base    & 88.6M & 0.841 & 0.899 \\
ConvNeXt-Large   & 197M & 0.836 & 0.882 \\
\bottomrule
\end{tabular}
\end{table}

\section{Conclusion}
\label{sec_conclusion}

In this work, we revisit the role of multimodal large language models in image quality assessment and argue that the main challenge lies not in perceptual understanding, but in aligning model outputs with dataset-specific MOS scales. Based on this insight, we propose LEAF, a label-efficient IQA framework that decouples perceptual knowledge transfer from MOS calibration. The MLLM teacher provides point-wise judgments and pair-wise preferences, while the student model learns robust quality perception without relying on human annotations in the distillation stage. Finally, a lightweight calibration stage with a small MOS subset is sufficient to align predictions with human annotations.

Extensive experiments on both user-generated and AI-generated IQA benchmarks demonstrate that LEAF significantly reduces the dependence on annotations while achieving competitive or superior correlation with human judgments. This work provides a practical alternative to direct regression with heavy annotation requirements. We believe the proposed framework offers a promising direction for future research on label-efficient quality assessment tasks.

\bibliographystyle{named}
\bibliography{ijcai26}

\end{document}